\documentclass[pdflatex,sn-basic]{sn-jnl}
\ifdefined\pdfinfoomitdate\pdfinfoomitdate=1\pdftrailerid{}\fi %
\ifdefined\pdfsuppressptexinfo\pdfsuppressptexinfo=-1\fi %

\usepackage[T1,T5]{fontenc}    %
\DeclareTextAccentDefault{\k}{T1}
\usepackage{graphicx}
\usepackage{booktabs}
\usepackage{multirow}
\usepackage{array}
\usepackage{amsmath,amssymb}
\usepackage{xcolor}
\usepackage{url}
\usepackage{xurl} %
\let\cite\citep
\renewcommand\labelitemi{\raise1pt\hbox{$\bullet$}}
\hypersetup{pdftitle={ViAdminShift: A Vietnamese Benchmark for Event-Level Knowledge Updates Built from the 2025
  Administrative Reform}, pdfauthor={Phan Binh Nguyen Lam and Nguyen Thai Anh}}

\newcommand{\lrefig}[2]{\includegraphics[width=#2]{#1}}
\newcommand{\pc}{\textsc{pc}}
\newcommand{\wc}{\textsc{wc}}
\newcommand{\cond}[1]{\texttt{#1}}
\newcommand{\nLocal}{26}
\newcommand{\nAPI}{41}
\newcommand{\nModels}{67}
\newcommand{\nKnowers}{20}
 \newcommand{\cue}[1]{\textsf{#1}}
\newcommand{\qt}[1]{\textsf{#1}}

\begin{document}

\title{ViAdminShift: A Vietnamese Benchmark for Event-Level Knowledge Updates Built from the 2025 Administrative Reform}

\author[1]{\sur{Phan} \fnm{Binh Nguyen Lam}}\email{24125014@student.hcmus.edu.vn}
\author*[2]{\sur{Nguyen} \fnm{Thai Anh}}\email{anh.nt@vlu.edu.vn}
\affil[1]{\orgdiv{Faculty of Information Technology}, \orgname{University of Science, Vietnam National University}, \orgaddress{\city{Ho Chi Minh City}, \country{Vietnam}}}
\affil*[2]{\orgdiv{Faculty of Information Technology, Van Lang School of Technology}, \orgname{Van Lang University}, \orgaddress{\city{Ho Chi Minh City}, \country{Vietnam}}}

\abstract{\unboldmath On July 1, 2025, Vietnam reduced its 63 provinces to 34, abolished its 696 districts, and replaced its 10,035 commune-level units (wards) with 3,321. \textsc{ViAdmin\-Shift}, a Vietnamese language resource about this event, consists of 7,995 benchmark items, 2,281 retrieval passages from pre- and post-reform Wikipedia revisions and dated templates, and old-to-new tables that a released script rebuilds from the official conversion table (not redistributed). Changed items store the pre-reform answer (except 121 address items), a $2{\times}2$ design uses untouched units as controls, and 400 addresses appear without diacritics. The tables agree with an independent lineage on 10,564 of 10,565 official pairs, disagreements were settled against the resolutions, and the probing-track scorer agrees with both authors' blind labels (Cohen's $\kappa=0.91$ and $0.89$). Data, outputs, code, and a datasheet are released. In the probing track, at least \nKnowers{} of 67 models know the reform. Fourteen of them give the new province alone for at most 11\% of plain questions about old addresses (old province only: 14--100\%); a bare date recovers up to 100\%. Post hoc (80 province-free addresses, single runs), with reasoning off, today's date in the system prompt leaves 68--100\% old-province-only answers for 10 of 12 models; with reasoning, an ``as of'' prefix in the question leaves none for seven of eight. On the 1,262-item benchmark subset, with same-source old and new passages, small models unaware of the reform give 14--30\% outdated answers and the five benchmark-track knowers (also differing in size or provider) 0--5\%. }


\maketitle

\section{Introduction}
\label{sec:intro}

\begin{figure}[t]
\centering
\lrefig{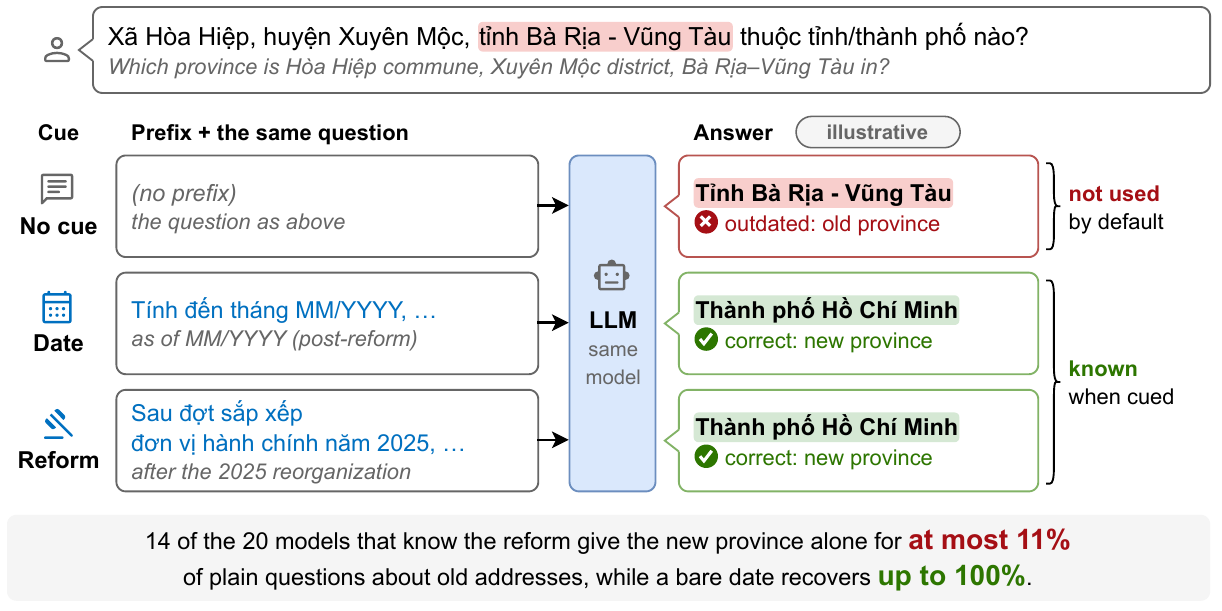}{\textwidth}
\caption{Most models that know the 2025 update do not use it by default. One model gets the same question with no cue, a bare post-reform date, or the reform named. Red marks the old province, green the new one, blue the cue prefixes, and gray italics the English glosses. The answers are illustrative; Fig.~\ref{fig:cue} gives the data, and the band the headline result (Section~\ref{sec:cue}).}
\label{fig:teaser}
\end{figure}

This paper presents \textsc{ViAdminShift}, a Vietnamese language resource built around one real event that changed thousands of facts at once, and reports what it reveals about large language models (LLMs). LLMs are trained on a fixed snapshot of text, while facts keep changing. Many resources sample isolated facts that changed for unrelated reasons, such as a soccer player's club or an office holder (Section~\ref{sec:related}). Real updates often arrive as \emph{events}, and a model that has absorbed an event should update every fact it entails. A model may also hold an update without using it, answering a plain question from the old map and a dated one from the new (Fig.~\ref{fig:teaser}).

Vietnam's 2025 administrative reform is such an event. Adopted on June 12, 2025, Resolution 202/2025/QH15 reduced the 63 provincial-level units to 34: 23 new units were formed by merging two or three old ones, and 11 were left unchanged~\cite{nq202_2025}. The Law on the Organization of Local Government left two levels, province and commune, and ended district governments on July 1, 2025~\cite{law72_2025}. At the commune level, the National Assembly Standing Committee approved 3,321 units in place of 10,035, in principle on June 5~\cite{vnplus_3321} and formally in Resolutions \mbox{1654--1687/NQ-UBTVQH15} of June 16, 2025. We call every commune-level unit a \emph{ward}: \emph{xã} (rural commune), \emph{phường} (urban ward), \emph{thị trấn} (township), or \emph{đặc khu} (special zone). Every ward of the 52 merged provinces now belongs to a new province, often under another name; almost all wards were merged, renamed, or divided; and every postal address lost its district segment: \emph{Phường Bến Nghé, Quận~1, Thành phố Hồ Chí Minh} is now \emph{Phường Sài Gòn, Thành phố Hồ Chí Minh}.

Three properties make the event a good basis for a language resource. Its date is exact, and it was adopted and reported weeks before the new units began to operate on July 1, 2025~\cite{vnplus_3321,nq202_2025}, so a model whose training data end in that window may know the headline without the details. Its consequences are specified by official tables, so gold answers are not a matter of judgment. And the units it left untouched form a built-in control group that separates outdated answers from other errors. The change also matters in practice: services must convert addresses between the two structures, open-source converters appeared soon after the reform~\cite{vietnamadminunits}, and pre- and post-reform documents coexist in retrieval stores. Translating an English temporal benchmark cannot produce such a resource: the sources exist only in Vietnamese, and the facts live in Vietnamese administrative naming, with changed unit-type words, recurring names, and omitted diacritics (Section~\ref{sec:vietnamese}).

We therefore build \textsc{ViAdminShift} from the official sources: deterministic templates, not an LLM, turn its canonical tables into questions with exact gold and pre-reform answers. The resource supports a sharper question than ``does the model know the new map?'': \emph{when a model knows the new map, does it use it?} We call the behavior in Fig.~\ref{fig:teaser} the \emph{temporal default}: the model holds both maps and, unless the question is anchored in time, answers from the old one. Log-probability comparisons~\cite{hossain2026right} and answers~\cite{zhao2024setclock} of pretrained models show a similar preference for older facts; we show it in the generated answers of deployed models, for an event-level change, with premise controls. Alongside strict accuracy (the new name alone), the prespecified headline metric, we report \emph{old-only} answers (Section~\ref{sec:metrics}), which give a user no sign of being outdated. Our contributions are:
\begin{itemize}
  \item \textbf{Canonical tables of the reform.} Tables of the units before and after the reform and 10,566 old-to-new ward pairs, derived from the official conversion table~\cite{nso_conversion}, checked against an independent source lineage and the resolutions, and rebuilt by a released script (Section~\ref{sec:sources}).
  \item \textbf{A benchmark with built-in controls.} Items of seven types, from headline facts to ward-level address conversion, that store the pre-reform answer where it differs (except 121 address items), place ward-level items in a $2{\times}2$ design (province changed $\times$ ward changed), and include diacritic-free address twins (Sections~\ref{sec:changetypes}--\ref{sec:vietnamese}). Eight local models (seven unaware of the reform) and four API models that know it ran all 7,995 items closed-book and, except three API models (1,460), with mixed passages.
  \item \textbf{Paired retrieval contexts.} Pre- and post-reform revisions of the same Vietnamese Wikipedia articles where they exist, and dated templates for the rest, which supply the new passage of 79\% of the items (Section~\ref{sec:contexts}).
  \item \textbf{Evaluation sets, scorers, and documentation.} Paired item sets that differ only in a temporal cue, a date's position, or a dropped address component; rule-based scorers with Vietnamese-specific normalization (the probing-track scorer validated against both authors' blind labels); and a datasheet~\cite{gebru2021datasheets}, a data statement~\cite{bender2018data}, and a versioning plan (Sections~\ref{sec:qc}--\ref{sec:release}).
  \item \textbf{What the resource reveals.} We probe 67 models (Section~\ref{sec:models}). Of the \nKnowers{} that know the reform, 14 give the new province alone for at most 11\% of plain questions about old addresses, and a bare date recovers up to 100\% (Section~\ref{sec:cue}). Post hoc, for most models, a date in the question leaves fewer old-only answers than the same date in the system prompt (Section~\ref{sec:anchor-res}). In the benchmark track, small models unaware of the reform give outdated answers with mixed passages far more often than the five that know it, which also differ in size or provider (Section~\ref{sec:rag}).
\end{itemize}

\section{Related Work}
\label{sec:related}

\subsection{Temporal and outdated knowledge}
\label{sec:rw-temporal}

Early work probed facts that change over time~\cite{dhingra2022time} and answers that depend on when a question is asked~\cite{zhang2021situatedqa}. TimeQA~\cite{chen2021timeqa} and TempReason~\cite{tan2023tempreason} ask for a Wikidata fact at a time stated in the question (or, in TempReason, relative to another event), TDBench~\cite{kim2026harnessing} generates time-sensitive questions from temporal databases, and other benchmarks follow change through consecutive Wikipedia and Wikidata snapshots~\cite{jang2022temporalwiki} or regularly refreshed questions~\cite{kasai2023realtime,vu2024freshllms}.

\begin{table}[t]
\caption{\textsc{ViAdminShift} and resources for temporal and outdated knowledge, as reported in TimeQA~\cite{chen2021timeqa}, TempReason~\cite{tan2023tempreason}, ChroKnowBench~\cite{park2025chroknowledge}, EvolvingQA~\cite{kim2024evolvingqa}, EvolveBench~\cite{zhu2025evolvebench}, evolveQA~\cite{nakshatri2025evolveqa}, WikiRecentChanges~\cite{wallat2026when}, PTC~\cite{hossain2026right}, FreshQA~\cite{vu2024freshllms}, and RT-QA~\cite{zhou2026benchmarking}. \emph{Ctrl.}: unchanged facts are scored with the changed ones. \emph{Old}: an answer correct before the change is scored as outdated for a later time. \emph{Present date}: where the model is told the current date. n/a: not stated or not verifiable. Licenses: TimeQA, BSD-3-Clause; TempReason, CC BY-SA 3.0; ChroKnowBench, CC BY 4.0; PTC, MIT (dataset card); FreshQA, Apache-2.0. The evolveQA data are not released.}
\label{tab:benchcompare}
\footnotesize
\renewcommand{\arraystretch}{1.15}
\setlength{\tabcolsep}{3pt}
\newcolumntype{L}[1]{>{\raggedright\arraybackslash}p{#1}}
\begin{tabular}{@{}L{49pt}L{20pt}L{56pt}L{104pt}L{22pt}L{30pt}L{55pt}@{}}
\toprule
Resource & Lang. & Size & Source and construction & Ctrl. & Old & Present date \\
\midrule
TimeQA & En & ${\sim}$20k per version & Crowd-verified Wikidata facts, template questions; year, month & No & No & None (time in question) \\
TempReason & En & 13,823 (test) & Wikidata 2022 and synthetic date arithmetic; month, year & No & Training only & None (time in question) \\
ChroKnow\-Bench & En & 37,606 variant, 27,373 invariant & Wikidata, UMLS, US federal regulations (variant); CSKG, Math-KG (invariant); yearly & Yes & No & None (year in query) \\
\midrule
EvolvingQA & En & 23,283 edited, 49,504 unchanged, 198,109 new & Monthly Wikipedia diffs (2023), QA written by GPT-3.5 & Yes & Yes & n/a \\
EvolveBench & En & 1,640 (132 subjects) & Wikidata histories, partly updated by hand; GPT-4o contexts; day & No & No & User prompt \\
evolveQA & En & 922 & AWS, Azure, and WHO update corpora; LLM-written questions & Yes & Yes & Hint before the cutoff \\
WikiRecent\-Changes & En & ${\sim}$2.4k changed, 2,381 stable & Wikidata changes of 2025, contexts by Qwen-3-14B & Yes & No & Prefix with date and cutoff (one variant) \\
PTC benchmark & En & 8,746 transitions & Wikidata, five position-holder relations, 2010--2024 & No & Yes & None (time in question) \\
\midrule
FreshQA & En & 600 & Human-written questions, mixed topics & Partly & n/a & Few-shot answers, with dated evidence (FreshPrompt) \\
RT-QA & Zh & 320 & Crawling workflows over live websites, checked by hand & No & n/a & System prompt \\
\midrule
\textsc{ViAdmin\-Shift} (ours) & Vi & 7,995 items, 10,566 mapping pairs (rebuilt by script) & Official table of the National Statistics Office (NSO), cross-checked; deterministic templates; real Wikipedia revisions and dated template passages; one legal event & Yes ($2{\times}2$) & Yes & None, system prompt, question, earlier turn, tool, chat template \\
\bottomrule
\end{tabular}
\end{table}

Closest to our design, EvolvingQA~\cite{kim2024evolvingqa} scores the outdated answer of each edited fact next to the updated one and adds unchanged facts, as our stored old answers and control units do, but GPT-3.5 writes its questions, and it tests the continual pretraining of T5 rather than the prompting of deployed LLMs. ChroKnowBench~\cite{park2025chroknowledge} contrasts changed facts with independently sampled facts that could have changed but did not; our controls are the units that the same event left untouched. WikiRecentChanges~\cite{wallat2026when} also pairs changed with stable facts. In evolveQA~\cite{nakshatri2025evolveqa}, models often pick the current answer in multiple choice but an outdated one in free-form answers. LLMs still give outdated answers~\cite{tang2025evowiki}, and outdated context misleads them~\cite{zhu2025evolvebench}.

\citet{hossain2026right} define \emph{parametric temporal conflict} (PTC): in four open-weight base models, the older position holder outscores the newer, still recoverable one for 4.1--10.3\% of the retained transitions, and a one-line date prefix restores the newer holder in 61--81\% of these. \textsc{ViAdminShift} instead covers one real, event-level update, with stored pre-reform answers and unchanged units as a built-in control.

\subsection{Where the present date is given}
\label{sec:rw-date}

Prior resources fix the present implicitly or state the date in fixed positions (Table~\ref{tab:benchcompare}). A date-and-cutoff prefix brings mostly small gains~\cite{wallat2026when}, and a date hint that precedes the model's cutoff changes accuracy little~\cite{nakshatri2025evolveqa}. \citet{sanzguerrero2026dating} show that the chat templates of Llama~3.1 and gpt-oss, and some API providers, insert the current date into the system prompt, and that changing only this date moves scores on time-independent benchmarks by up to 14 points. \citet{herel2025time} compare wordings of a date prefix in one position, and in RT-QA~\cite{zhou2026benchmarking}, search agents given the timestamp in the system prompt re-anchor to past dates that they retrieve. Placement matters for other prompt content, such as the answer instruction of an Italian benchmark~\cite{musacchio2026italian}; we compare system-prompt, in-question, earlier-turn, and tool placements of the same date on time-sensitive facts (post hoc; Section~\ref{sec:anchor-res}).

\subsection{Knowledge conflicts, outdated retrieval, and editing}
\label{sec:rw-conflicts}
\label{sec:rw-editing}

When retrieved context contradicts parametric memory, fine-tuned QA readers overrely on memorized answers under entity substitution~\cite{longpre2021entity}, while LLMs accept coherent counterevidence but, given both supporting and conflicting evidence, often side with memory~\cite{xie2024adaptive}. Outdated documents in a retrieval store distract models from current information~\cite{ouyang2025hoh}, and several versions of a fact in context mislead models~\cite{ozer2025question}. Many conflict sets use substituted, constructed, or LLM-generated contexts~\cite{longpre2021entity,xie2024adaptive,wallat2026when}, whereas each of our \emph{mixed} passages was true at its date: a real Wikipedia revision or, without an article, a dated template (Section~\ref{sec:contexts}).

Knowledge-editing benchmarks ask whether an injected fact spreads to entailed and multi-hop facts~\cite{cohen2024evaluating}, and work on event-level editing~\cite{peng2024event} argues that real updates arrive as events that change many triples at once. Many of these edits are counterfactual or constructed, and even real-change subsets, such as the Recent subset of RippleEdits~\cite{cohen2024evaluating}, inject each change as an independent triple. The reform is a real event-level ``edit'': we edit nothing and ask whether a model that knows the headline carries it to the entailed ward-level facts.

\subsection{Vietnamese and Southeast Asian evaluation}
\label{sec:rw-vi}

Vietnamese LLM benchmarks cover school exams with geography questions~\cite{dao2023vnhsge}, general multitask understanding~\cite{nguyen2024villm,bui2025vmlu}, and law. VLegal-Bench~\cite{dong2025vlegal} reports ``recency blindness'' in article prediction (models pick outdated legislation over 2024--2025 instruments) and includes one task on amendment and replacement relations between legal documents, but none of these benchmarks is built around a real-world knowledge update. AC-IQuAD~\cite{doxolodeo2024aciquad} generates Indonesian questions from Wikidata triples and reports that low-content, probably bot-created Wikipedia articles on districts skewed its questions toward administrative units; most Vietnamese wards have no article, so many of our passages are generated from the tables (Section~\ref{sec:contexts}).

\section{The ViAdminShift Resource}
\label{sec:benchmark}

\begin{figure}[t]
\centering
\lrefig{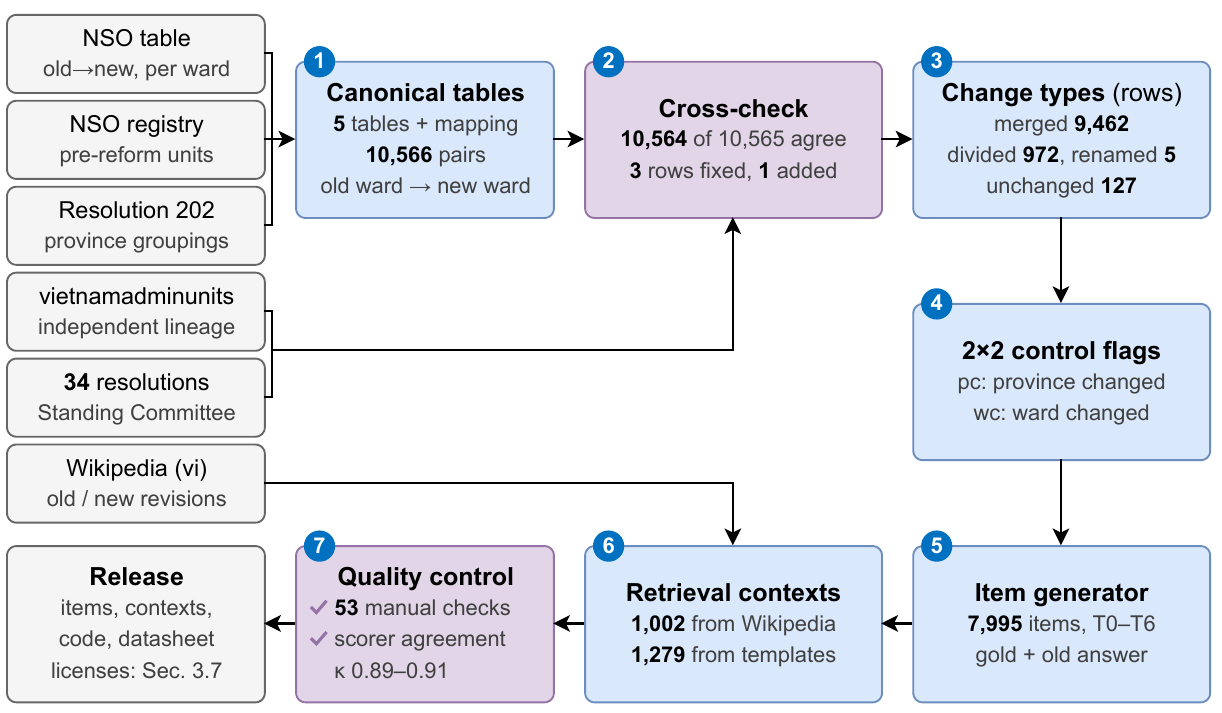}{\textwidth}
\caption{\textsc{ViAdminShift} is built from official sources (Wikipedia only for the retrieval contexts), cross-checked, and released with its documentation; the numbers are those of Sections~\ref{sec:sources}--\ref{sec:release}.}
\label{fig:pipeline}
\end{figure}

\subsection{Sources and canonical tables}
\label{sec:sources}

\begin{figure}[t]
\centering
\lrefig{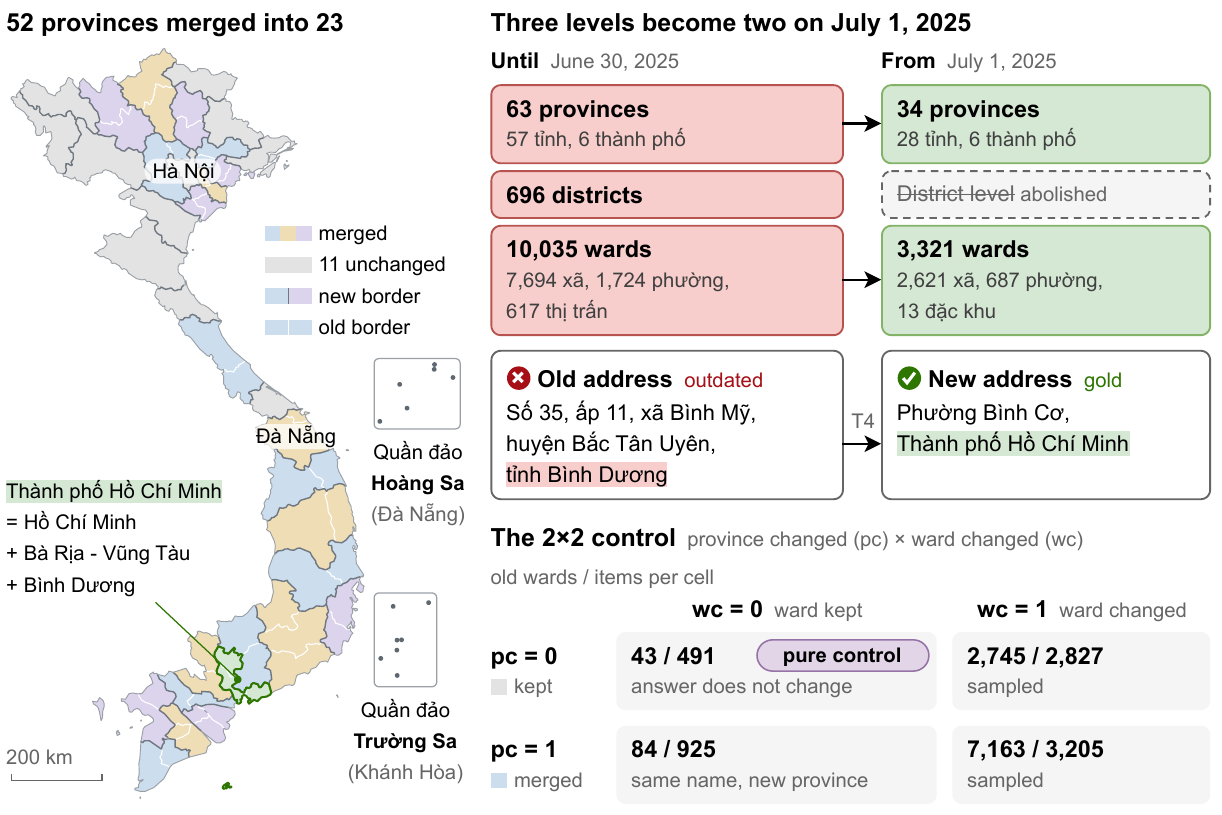}{\textwidth}
\caption{The reform merged 52 of the 63 provinces into 23 new ones and removed the district level, and the $2{\times}2$ control separates outdated answers from other errors. Left: the 34 provinces since July 1, 2025~\cite{nq202_2025}; merged provinces have pastel fills with their old borders in white, unchanged ones are gray, and the running example of Fig.~\ref{fig:teaser} is green. Top right: the structure on June 30, 2025 (red) and on July 1, 2025 (green), with pre-reform unit types as officially reported~\cite{moha2025units}, and one \qt{T4} item (real unit names; synthetic house number and hamlet). Bottom right: old wards in the full mapping / benchmark items per cell (all ward-level item types, before removing the 779 copyable items of Section~\ref{sec:items}), crossing whether the old province was merged (\pc{}) with whether the ward changed (\wc{}); the pure control is purple. The 547 items about provinces, districts, or headline facts have no cell. Boundaries: geoBoundaries (public domain; \citealp{runfola2020geoboundaries}), simplified; Hoàng Sa and Trường Sa are shown schematically.}
\label{fig:reform}
\end{figure}

The backbone of \textsc{ViAdminShift} (Fig.~\ref{fig:pipeline}) is the official old-to-new conversion table~\cite{nso_conversion}, which lists the old wards of every new ward. It is published by the National Statistics Office (NSO, Cục Thống kê). The pre-reform units come from the same office's registry, as distributed with vietnamadminunits~\cite{vietnamadminunits}; all its 10,035 wards occur in the NSO table. Province-level groupings were transcribed from Resolution 202/2025/QH15~\cite{nq202_2025}, and the build aborts unless the NSO table implies exactly these groupings. We derive five canonical tables, with 63 provinces, 696 districts, and 10,035 wards before the reform and 34 provinces and 3,321 wards after it (Fig.~\ref{fig:reform}, top right), and a mapping with one row for each of the 10,566 (old ward, new ward) pairs.

For cross-checking we use vietnamadminunits~\cite{vietnamadminunits}, a mapping built independently from the composition texts of the government's merger portal. Two further public mappings copy these lineages: undertheseanlp/\allowbreak{}address~\cite{undertheseanlp_address} copies vietnamadminunits, and vietaddr~\cite{vietaddr} the NSO table. The two lineages agree on 10,564 of the 10,565 pairs in the NSO table. We resolved every disagreement against the 34 ward-level resolutions of the National Assembly Standing Committee: we read the relevant clauses in all but five same-name cases, which an automated comparison of every clause with the mapping confirmed. The NSO table was right except for three wrong old-ward codes (corrected) and one omitted pair (added), a ward split between two new wards. The other 32 pairs only in vietnamadminunits contradict the resolutions (mostly same-name confusions).

\subsection{Change types and the \texorpdfstring{$2{\times}2$}{2\texttimes 2} control}
\label{sec:changetypes}
\label{sec:control}

Each mapping row from old ward $o$ to new ward $n$ gets the first applicable change type: \textbf{divided} if $o$ maps to more than one new ward (972 rows, 441 old wards), \textbf{merged} if $n$ is formed from more than one old ward (9,462 rows, including 1,641 in which $n$ keeps $o$'s name), \textbf{unchanged} if the mapping is one-to-one and name and unit type are kept (127), and \textbf{renamed} otherwise (5). Because divided wards have no unique successor, they appear only in items that ask for the new province.

Each ward-level item carries two flags (Fig.~\ref{fig:reform}, bottom right): \pc{} (\emph{province changed}), true when the old province was merged, even if the new province kept its name, and \wc{} (\emph{ward changed}), true unless the change type is \emph{unchanged}. In the pure control ($\pc{}=0,\wc{}=0$) the answer did not change (\emph{Xã Mù Cả} in Mường Tè, Lai Châu), so an outdated answer about the reform there can only be a \qt{T4} address in the pre-reform three-level form. In the cell $\pc{}=1,\wc{}=0$, a ward keeps its name but changes province (\emph{Xã Ngọc Long} in Yên Minh, Hà Giang, now in Tỉnh Tuyên Quang).

\begin{table}[t]
\centering
\caption{Item types with one real (cue-free) item each and the number of items. Multiple-choice (MCQ) items have four options: the gold, the outdated answer if one exists, and the nearest new provinces (\qt{T1}, \qt{T6}) or new wards of the same new province (\qt{T5}).}
\label{tab:qtypes}
\footnotesize
\setlength{\tabcolsep}{4pt}
\begin{tabular}{@{}>{\raggedright\arraybackslash}p{1.75cm}>{\raggedright\arraybackslash}p{6.45cm}>{\raggedright\arraybackslash}p{3.15cm}r@{}}
\toprule
Type & Example (Vietnamese; \emph{English gloss}) & Gold / outdated answer & Items \\
\midrule
\qt{T0} headline (MCQ, short) &
Cả nước hiện có tổng cộng bao nhiêu tỉnh và thành phố trực thuộc trung ương?
\emph{How many provinces and centrally run cities does the country have now?} &
34 / 63 & 12 \\
\qt{T6} reverse (MCQ, short) &
Địa bàn tỉnh Bình Dương thuộc đơn vị hành chính cấp tỉnh nào?
\emph{Which provincial-level unit is the area of Bình Dương province in?} &
Thành phố Hồ Chí Minh / Tỉnh Bình Dương & 378 \\
\qt{T2} composition (MCQ) &
Thành phố Hồ Chí Minh hiện nay được hình thành từ (những) đơn vị hành chính cấp tỉnh nào trước đây?
\emph{Which former provincial units is Ho Chi Minh City (HCMC) now formed from?} &
HCMC, Bà Rịa--Vũng Tàu, and Bình Dương / HCMC only & 34 \\
\qt{T1} membership (MCQ) &
Xã Hòa Hiệp, huyện Xuyên Mộc, tỉnh Bà Rịa - Vũng Tàu thuộc tỉnh/thành phố nào?
\emph{Which province is Hòa Hiệp commune, Xuyên Mộc district, Bà Rịa--Vũng Tàu in?} &
Thành phố Hồ Chí Minh / Tỉnh Bà Rịa - Vũng Tàu & 2,781 \\
\qt{T1} new name (MCQ) &
Phường Thuận Giao thuộc tỉnh/thành phố nào?
\emph{Which province is (the new) Thuận Giao ward in?} &
Thành phố Hồ Chí Minh / Tỉnh Bình Dương & 720 \\
\qt{T3} existence (MCQ) &
Hiện nay, tỉnh Bình Dương có phải là một đơn vị hành chính đang tồn tại không? Nếu có, đó là đơn vị hành chính cấp nào?
\emph{Is Bình Dương province an existing administrative unit today? If so, of which level?} &
No / yes, provincial level & 308 \\
\qt{T4} address (short) &
Địa chỉ cũ: Số 35, ấp 11, xã Bình Mỹ, huyện Bắc Tân Uyên, tỉnh Bình Dương. Địa chỉ này hiện nay được viết theo đơn vị hành chính hai cấp (cấp xã, cấp tỉnh) như thế nào?
\emph{Old address: No.~35, hamlet 11, Bình Mỹ commune, Bắc Tân Uyên district, Bình Dương. How is it written in the two-level structure?} &
Phường Bình Cơ, Thành phố Hồ Chí Minh / (old address) & 1,200 \\
\qt{T5} successor (MCQ, short) &
Địa bàn xã Lạc An, huyện Bắc Tân Uyên, tỉnh Bình Dương thuộc đơn vị hành chính cấp xã nào?
\emph{Which commune-level unit is the area of Lạc An commune (Bắc Tân Uyên, Bình Dương) in?} &
Xã Thường Tân / Xã Lạc An & 2,562 \\
\midrule
Total & 5,320 MCQ and 2,675 short-answer items, among them 400 diacritic-free \qt{T4} twins & & 7,995 \\
\bottomrule
\end{tabular}
\end{table}

\subsection{Item types}
\label{sec:items}

Table~\ref{tab:qtypes} lists the seven item types: \qt{T0}, \qt{T2}, and \qt{T6} ask about headline and province-level facts, and \qt{T1}, \qt{T4}, \qt{T5}, and part of \qt{T3} about the facts that the reform entails for individual wards. \qt{T1}, \qt{T5}, and \qt{T6} items come in three cue variants that differ only in a prefix (Section~\ref{sec:ladder}). Items whose answer changed also store the pre-reform answer (\texttt{old\_answer}), which multiple-choice items include as a distractor, so a wrong answer can be labeled \emph{outdated}. The field is empty where the two answers coincide, for 121 \qt{T4} items whose ward and province names did not change, and for 87 new-name \qt{T1} items about wards whose name did not exist before the reform (2,989 items in all).

The control cells ($\wc{}=0$) are used in full, and the two changed cells are sampled round-robin over old provinces, using only old wards whose names changed: for $\pc{}=1$ the province name, and for $\wc{}=1$ the bare name of every successor. Ward-level items (\qt{T1}, \qt{T3}--\qt{T5}) cover 1,708 of the 10,035 old wards and 57 of the 63 old provinces. In 849 \qt{T4} and \qt{T5} items with a changed answer (109 of the 1,262-item subset of Section~\ref{sec:tracks}), the new ward's bare name occurs in the question, compared without diacritics (118 of them are diacritic-free \qt{T4} twins). In 779 of them (104 in the subset), the ward was renamed, so the answer can be copied; the other 70 are \qt{T4} addresses of wards that kept their name, where only the province changed and nothing is leaked. We remove the 779 items from the $2{\times}2$ analysis and the released per-type tables. Elsewhere they are kept, and removing them changes accuracy by at most 4.1 percentage points closed-book and 4.8 points with a retrieved context, for any model and condition (all 849: 4.1 and 5.4).

\subsection{Vietnamese-specific properties}
\label{sec:vietnamese}

Counts in this section come from the undertheseanlp/\allowbreak{}address copy of the independent lineage.

\textbf{Naming.} White space separates every syllable, so it does not mark word boundaries~\cite{nguyen2017ensuring}. Both scorers therefore strip the unit-type word before each name (the reform changed this word for some units) and match names as sequences of whole syllables. The benchmark-track scorer also compares that word by level and never accepts \emph{thành phố} (a city or a district) for a ward. Names recur: 43\% of pre-reform wards share their bare name with another (\emph{Tân Thành} occurs 34 times), so every item about an old ward names its old district and province.

\textbf{Spelling.} Users often type Vietnamese without diacritics, which makes text ambiguous~\cite{dang2020tdp}: without them, 17.5\% of pre-reform wards share their name with a differently spelled ward (\emph{Tân Thanh}, \emph{Tân Thành}, \emph{Tân Thạnh}). Half of the 800 \qt{T4} addresses therefore also appear as diacritic-free twins, which the benchmark-track scorer matches without diacritics, mapping \emph{đ} to \emph{d}. Tone marks in \emph{oa}, \emph{oe}, and \emph{uy} follow one of two conventions (\emph{Hòa} or \emph{Hoà}) that Unicode normalization (NFC) does not unify. Both scorers unify them (and dashes), which changes 147 of 121,324 cue-ladder labels (post hoc).

\subsection{Retrieval contexts}
\label{sec:contexts}

For each province and district article of Vietnamese Wikipedia, we take the last revision before June 12, 2025, that does not yet describe the reform, and the last revision before January 1, 2026, if it postdates July 15, 2025. Passages are excerpts of at most 1,250 characters from the lead or the administrative section. Most wards have no article, so the other items use template passages generated from the canonical tables in the same format, dated June 1 or October 1, 2025, and citing the relevant resolution. The 2,281 passages used are 1,002 Wikipedia excerpts (575 old, 427 new) and 1,279 templates (277 old, 1,002 new). The \cond{new} condition shows the relevant new passage(s), and the \cond{mixed} condition old and new passages in a fixed shuffled order.

\subsection{Quality control}
\label{sec:qc}

The adjudication of Section~\ref{sec:sources} used HTML reproductions of the resolutions, not the official scans; the manual checks used the official texts. The authors found no error in 53 manual checks: 30 random mapping rows (seed 13, from 30 different new provinces) with the full composition of their new wards; 11 targeted checks of divided, conflicting, or corrected units; and the gold answers of the 12 \qt{T0} items, checked against Resolution 202/2025/QH15. The authors read every cited clause and made each judgment themselves; the checklist was prepared with Claude's help. Audit code written separately from the generator (both with Claude's help) rederived the gold and old answers of the other 7,983 items and found no gold error; a linter found no old answer scored as correct and no malformed options.

The two authors, both native speakers, labeled 200 probing-track answers to absorbed \qt{T1} and \qt{T1np} units (Section~\ref{sec:ladder}; 40 per scorer label when sampled, \emph{both} split by name order) independently and blind: their sheets showed neither the scorer's label nor the model and listed the rows in different orders. They agree with each other on 97\% ($\kappa=0.96$) and with the scorer on 93\% and 92\% ($\kappa=0.91$ and $0.89$). Most disagreements (13 of 14 and 15 of 17) go against the scoring rules, which the authors know: for an answer that names both provinces, the scorer takes the name given first and the guideline the province the answer commits to. The scorer thus undercounts new-first answers: the annotators read about a third of its sampled old-first answers as new-first, among them long explanations that end on the new province (including all four sampled from Claude Opus~4.8). Besides the copyable items, two item issues remain: 19 \qt{T3} items about new wards do not separate outdated from updated models, and three \qt{T1} items about \emph{Xã Đồng Nai} name its new province.

\subsection{Release, versioning, and maintenance}
\label{sec:release}

The release contains the items, the retrieval contexts, the evaluation sets of Section~\ref{sec:setup} (with the exact prompts), the model outputs behind every result, the scorers, the analysis scripts, and a datasheet~\cite{gebru2021datasheets}. Data, model outputs, and annotation sheets are released under CC~BY-SA~4.0 (required for the Wikipedia passages; outputs of tiny-aya-global: CC~BY-NC-SA~4.0), and the code under the MIT license; users of hosted-model outputs must also follow the providers' terms. The items contain only unit names, old-to-new assignments, and official unit codes, facts set by public legal acts. The NSO file carries no explicit license; no source file with NSO or government-portal data is redistributed, nor are the canonical tables, which reproduce the NSO file as a whole; the released code downloads these files and, if they match the versions used here, rebuilds the tables byte for byte.

Version 2025-07-01 describes the structure in force on July 1, 2025 (``old'' is the structure on June 30, 2025). Items about the present (``now,'' \emph{hiện nay}) refer to this date, and gold answers under the later date cue assume that no name or membership changed in between. Later changes we know of concern unit type only. Between April 30 and September 20, 2026, Đồng Nai, Quảng Ninh, and Bắc Ninh became centrally run cities with the same name and territory (Resolutions 30, 36, and 39/2026/QH16), and 10 xã of Đồng Nai and 12 of Bắc Ninh became phường under the same names (Resolutions 237 and 388/NQ-UBTVQH16). The date in the ``as of'' cue and in the ``today is'' sentence (Section~\ref{sec:ladder}) falls after the 2025 reform and after all of these 2026 changes. Because the scorers strip unit-type words and options differ by name, only \qt{T3} items, which name the unit type, are affected: four of them (\texttt{T3-prov-22}, \texttt{-27}, \texttt{-75}, and \texttt{T3-dist-256}) are correct only at the version date, and none about a ward names one of the 22 communes that became phường. Later versions will be rebuilt from then-current official sources, and released versions will remain unchanged. The resource is maintained by the corresponding author; errors can be reported on the repository's issue tracker, and confirmed corrections are listed in the changelog.

\section{Experimental Setup}
\label{sec:setup}

\subsection{Two evaluation tracks}
\label{sec:tracks}

\begin{figure}[t]
\centering
\lrefig{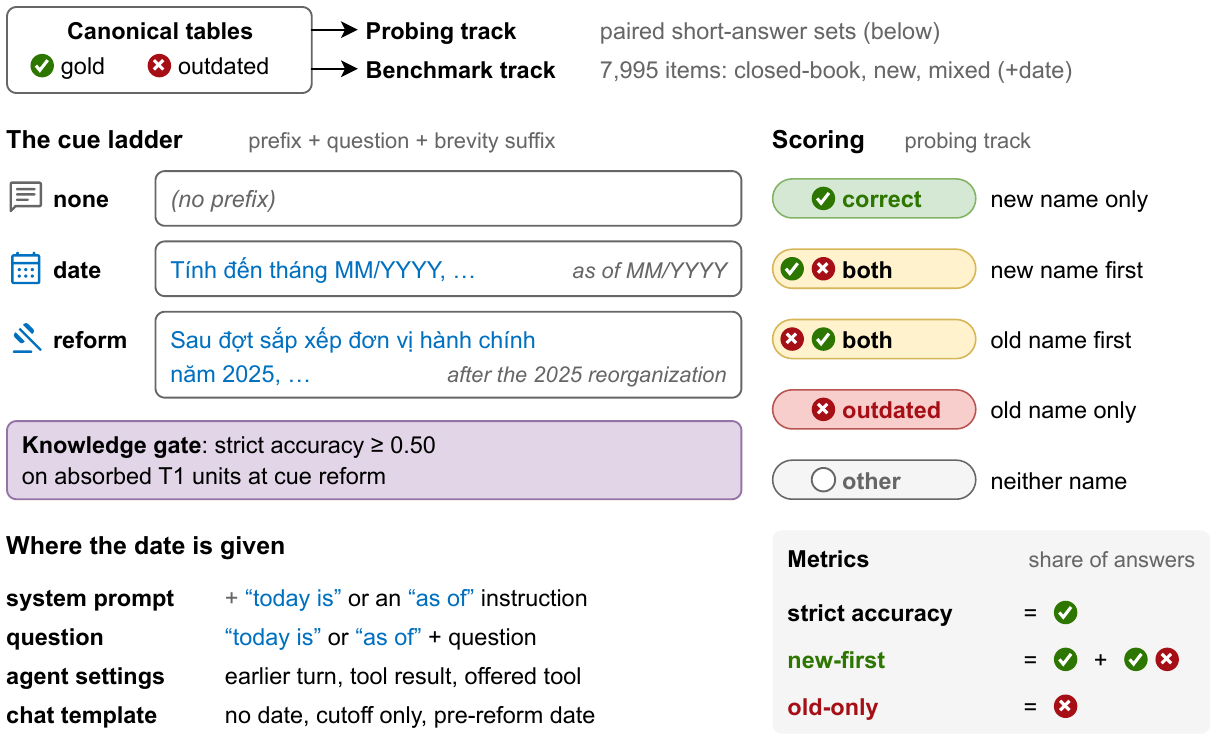}{\textwidth}
\caption{Evaluation design. Top: the gold and outdated answers that both tracks share, and the benchmark track's conditions. Left: the probing track's cue ladder, its knowledge gate (purple), and where a date can be given. Right: how probing-track answers are labeled and counted as strict accuracy (prespecified), new-first, and old-only (the last two post hoc). Green, new name; red, old name; yellow, both names; blue, the added cue or date; gray italics, English glosses.}
\label{fig:protocol}
\end{figure}

Both tracks take gold and outdated answers from the canonical tables (Section~\ref{sec:sources}; Fig.~\ref{fig:protocol}, top). The \emph{probing track} uses small, paired short-answer sets whose questions exist in versions that differ only in a cue, a system prompt, or one dropped component, so comparisons are paired by unit. The \emph{benchmark track} runs all 7,995 released items (Section~\ref{sec:items}) or a seeded 1,262-item subset stratified by item type and format, covering every $2{\times}2$ cell, closed-book and under the \cond{new} and \cond{mixed} conditions of Section~\ref{sec:contexts}, the latter also with dated passage headers (+date). Eight local models ran it (seven small ones of 1.7B to 4B unaware of the reform, and Muse-Glimmer-30B, which ran \cond{new} and +date only on the subset), as did four API models aware of it: \mbox{DeepSeek-V4.1}-Flash on all items in all conditions, and GPT-6-Luna, Claude Haiku~5.5, and Sonnet~5.5, which ran with passages only on the subset and the 220 items with a new Wikipedia excerpt and an old template passage (22 of them in the subset; 1,460 in all).

\subsection{Models and inference}
\label{sec:models}
\label{sec:inference}

\begin{table}[t]
\caption{Evaluated models by access route (number in parentheses). Local model names state the size, except Ling-3.0-tiny; most API models do not disclose it. The last column gives the requested setting and, for three routes, the measured share of cue-ladder items with reasoning (Section~\ref{sec:inference}).}
\label{tab:models}
\scriptsize
\setlength{\tabcolsep}{3pt}
\begin{tabular}{@{}>{\raggedright\arraybackslash}p{1.85cm} >{\raggedright\arraybackslash}p{6.24cm}
>{\raggedright\arraybackslash}p{4.51cm}@{}}
\toprule
Route & Models & Reasoning setting \\
\midrule
DeepSeek API (2) & DeepSeek-V4.1-Flash, DeepSeek-V4-Pro & \texttt{thinking: \{type: disabled\}} \\
\addlinespace
Anthropic API, OpenAI-compatible endpoint (13) & Claude Haiku~4.5, Opus~4.5, Sonnet~4.6, Opus~4.6, Opus~4.7, Opus~4.8,
Sonnet~5, Opus~5, Fable~5, Fable~5.1, Haiku~5.5, Sonnet~5.5, Opus~5.5; Sonnet~5.5 (thinking off via
\texttt{between\_tools}), Haiku~5.5, and Fable~5.1 also run at a second setting & thinking disabled (measured: Opus~4.8 38\%,
Haiku~5.5 22\%, Sonnet~5.5 15\%, others $\le$8\%); \texttt{reasoning\_effort: none} for Opus~5.5 (100\%),
Fable~5 (75\%), and Fable~5.1 (82\%) \\
\addlinespace
OmniRoute reseller (20) & GPT-6, GPT-5.5, gpt-oss-120B, Kimi-K3, Kimi-K2.5, GLM-5.3, GLM-5.3-Flash, Gemini-3.1-Pro,
Gemini-3.8-Flash, \mbox{Gemma-4-31B}, MiMo-v2.6-Flash, MiMo-v2.6-Pro, MiniMax-M2.7, MiniMax-M3, Muse-Spark-1.2$^*$,
Qwen3.6-27B, Qwen3.8-27B, Qwen3.8-Flash$^*$, Qwen3.8-Max, Qwen3.8-2.4T$^*$ & thinking disabled in the request, yet
reasoning on 79--100\% of items for GPT-6, GPT-5.5, Kimi-K3, both GLM-5.3, Gemini-3.1-Pro, gpt-oss-120B,
MiniMax-M2.7, Qwen3.6-27B, and the invalid runs ($^*$), and on at most 50\% for the rest \\
\addlinespace
OpenRouter, default provider routing (7) & GPT-6-Luna, GPT-5.6-Terra, Tencent HY4 (preview), Grok-4.7,
Gemini-4-Argon, LongCat-2.0, Mistral-Large-4; reruns of GPT-6 and Kimi-K3, the benchmark run of Claude
Haiku~5.5; not counted: Step-5 (preview),
stopped for budget after 117 of the 549 base items (gate: 68 of 80 correct) & \texttt{reasoning: \{enabled: false\}},
or \texttt{true} for the reasoning-on runs; \texttt{\{effort: low\}} where reasoning cannot be disabled (Grok-4.7,
Gemini-4-Argon) \\
\addlinespace
Local, llama.cpp server on GPU or CPU (26) & Muse-Glimmer-30B; Qwen3-1.7B, 4B, and 8B; Qwen3.5-2B, 4B, and 9B;
Qwen3.5-9B-Kimi-K3-Distilled; Qwen3.6-35B-A3B; Qwen3.8-27B; Qwen-SEA-LION-v4.5-27B; Nemotron-SEA-LION-v4.8-30B-A3B;
\mbox{NVIDIA-Nemotron-3.5-Lightning-30B-A3B}; SeaLLMs-v3-7B; \mbox{MiMo-V2.6-Distill-Qwen-9B}; Gemma-3-4B; Gemma-4-E4B;
Llama-3.2-3B; Ministral-3-8B; Granite-4.2-3B and 8B; LFM2.5-8B-A1B; Ling-3.0-tiny; \mbox{MiniCPM5-2B}; gpt-oss-20B;
tiny-aya-global~(3.35B) & \texttt{enable\_thinking: false} in the chat template (CPU runs also
\texttt{reasoning\_effort: low}); Muse-Glimmer at \texttt{reasoning\_strength: none}; Muse-Glimmer, gpt-oss-20B,
and LFM2.5-8B-A1B still reason on every cue-ladder item \\
\bottomrule
\end{tabular}
\end{table}

We evaluate \nModels{} distinct models, \nLocal{} local and \nAPI{} only through APIs, with \mbox{Qwen3.8-27B} on both routes (Table~\ref{tab:models}); reruns for route and reasoning comparisons are not counted. Local models use Q8\_0 GGUF weights, except gpt-oss-20B (MXFP4) and, in the probing track only, nine that we converted to bf16. The reseller OmniRoute does not name its upstream provider; with no cue, Claude Opus~5.5 answers 41 of the same 80 absorbed \qt{T1} units correctly through the first-party API and 44 through the reseller, and Qwen3.8-27B fails the gate on both routes.

Decoding is greedy (temperature 0) where the API accepts it (not for Claude models from Opus~4.7 on), and reasoning is set as low as each route allows (Table~\ref{tab:models}). Because API flags do not reveal whether a model reasoned, we \emph{measure} it: an item counts as reasoned if the API returns reasoning text or the reported completion tokens exceed the visible length (UTF-8 bytes/3) by more than 40, which holds for 3,056 of the 3,057 cue-ladder items of Claude Opus~5.5 and for no placement item of \mbox{DeepSeek-V4.1}-Flash. The measure also shows reasoning on most reseller models despite the disabled flag (Table~\ref{tab:models}); the knowers (Section~\ref{sec:prereg}) GPT-6, \mbox{GPT-5.5}, \mbox{Kimi-K3}, and \mbox{GLM-5.3} reasoned on 97--100\% of absorbed items with today's date in the system prompt. GPT-6 and Kimi-K3 were therefore rerun via OpenRouter with reasoning off and on for the date experiments; the other reseller models were not rerun. Benchmark-track answers are capped at 16/64 tokens (MCQ/short; Muse-Glimmer 1,024, without its template date) and scored even if truncated. Claude answers that hit the cap were re-asked at 64/128, then 512 (Sonnet started at 64/128; all cut answers but Haiku's 296 at 64/128 are released); 85 Sonnet answers remain truncated. Counting as invalid Sonnet's 221 subset answers longer than any uncut answer of Haiku's 16/64 pass (12 UTF-8 bytes for MCQ, 153 for short) lowers its closed-book subset accuracy from 71\% to 63\%. Sonnet ran as in its no-thinking probe, Haiku and \mbox{GPT-6-Luna} via OpenRouter (temperature 0 requested, reasoning off, provider not recorded; Haiku's gate probe on the first-party API); reasoning was not measured.

Each cell is a single run (repeats for four models: Section~\ref{sec:limitations}). Hosted models can change behind a fixed identifier, need not be deterministic under greedy decoding~\cite{atil2024nondeterminism}, and may add the current date~\cite{sanzguerrero2026dating}, so the released outputs are the reference for every number. Only the reseller was probed for a provider-added date: GPT-5.5 knew today's date in all seven probes, so its no-cue conditions are not date-free; GPT-6, Kimi-K3, and both GLM-5.3 models did not (one-off recitations of a dated system prompt by Kimi-K3 and GLM-5.3 did not recur).

\subsection{The cue ladder and date placement}
\label{sec:ladder}
\label{sec:anchor}

Each question is asked at three \emph{cue} levels that differ only in a prefix (Fig.~\ref{fig:protocol}, left): \cue{none}; \cue{date}, ``Tính đến tháng $\langle$MM/YYYY$\rangle$, \ldots'' (\emph{as of $\langle$month year$\rangle$}); and \cue{reform}, ``Sau đợt sắp xếp đơn vị hành chính năm 2025, \ldots'' (\emph{after the 2025 reorganization}). Here and below, the date is a fixed post-reform date close to the time of the runs, the same in every prompt except the pre-reform controls; the exact strings are in the released item files. Every prompt ends with a brevity suffix (\emph{answer briefly with the name only; do not explain}). Probes are short-answer versions of the \qt{T6}, \qt{T1}, and \qt{T5} items of Table~\ref{tab:qtypes}. In \qt{T6} and \qt{T1}, absorbed units (the old province merged into a differently named one) are the test and kept units the control; \qt{T1np} drops the province and uses only ward--district pairs unique nationwide, so the old province must be recalled, not copied. In \qt{T5}, renamed wards are the test and unrenamed ones the control, excluding items whose new ward name occurs in the old address (ignoring tone placement post hoc). The base set has 549 items: the 63 old provinces in \qt{T6} (29 absorbed, 34 kept) and 120 \qt{T1} units (80 absorbed, 40 kept). Knowers also answer 360 \qt{T1np} items (120 units), some models a 2,589-item extension (63 provinces, 400 \qt{T1} and 400 \qt{T5} units), and the two DeepSeek models and Qwen3-4B its superset with 1,000 absorbed \qt{T1} units.

\textbf{Date placement (post hoc).} On the \cue{none} questions of \qt{T6}, \qt{T1}, and \qt{T1np} (303 items), a plain system prompt (\emph{you are a helpful assistant}) is compared with the same prompt plus ``Hôm nay là ngày $\langle$DD/MM/YYYY$\rangle$.'' (\emph{today is $\langle$date$\rangle$}). This ``today is'' sentence and the ``as of'' date each appear in the system prompt (there as an instruction) or in the question (four cells), and two controls give a pre-reform date (January 1, 2025, as today's date in the system prompt, or ``as of January 2025'' in the question). Six models with reasoning off repeat the four cells on 300 absorbed \qt{T1np} units from all 29 absorbed provinces. \emph{Agent settings} give the date in an earlier turn, as a prior \texttt{get\_current\_date} result, or through an offered tool (with or without a call instruction), or let DeepSeek-V4.1-Flash, told the date, rewrite each question first. In Muse-Glimmer's chat template, the block stating its cutoff and today's date becomes no date, the cutoff only, or a pre-reform date. Further controls use post-reform ward names (240 wards), ``hiện nay'' (\emph{currently}) or a content-free filler as prefix, no brevity suffix (three models), and three repeats of the 80 absorbed \qt{T1} units (four models).

\subsection{Prespecified criteria}
\label{sec:prereg}

After a pilot (284 items, DeepSeek-V4.1-Flash) had suggested the effect, and before scoring the final ladder, we recorded these criteria in writing. A model \textbf{knows the update} (a \emph{knower}) if its strict accuracy on absorbed \qt{T1} units at \cue{reform} is at least 0.50 (the \emph{knowledge gate}). Its knowledge is \textbf{cue-dependent} if \cue{reform}$-$\cue{none} on those units is at least 0.30 with a paired bootstrap 95\% confidence interval excluding 0; as the gate requires 0.50 at \cue{reform}, a knower can miss this only with a no-cue accuracy above 0.20. \textbf{Premise-leak control}: in models below the gate, \cue{reform}$-$\cue{none} on absorbed units must stay below 0.15, and in all models $|$\cue{reform}$-$\cue{none}$|$ on kept units below 0.10; otherwise the cue effect is reported as partly premise-induced. \textbf{Generality} requires at least two model families that know the update to be cue-dependent. A validity rule in the gate script, not part of the recorded criteria, marks runs (one model and setting on one item set) with more than 20\% empty outputs as \emph{invalid} rather than negative. All other analyses, including every date-placement, agent, and template experiment, are post hoc and labeled as such in Section~\ref{sec:results}; benchmark-track analyses are descriptive.

\subsection{Scoring}
\label{sec:metrics}

The probing-track scorer (Fig.~\ref{fig:protocol}, right; validated in Section~\ref{sec:qc}) removes reasoning text and unit-type words (Section~\ref{sec:vietnamese}), normalizes answers (Unicode NFC, lowercase), and matches them as whole names against the gold and old names, with a small alias list. An answer is \emph{correct} if it names only the new unit, \emph{outdated} if only the old one, \emph{both} if both, and \emph{other} if neither (wrong unit, refusal, or empty). Strict accuracy (correct only) is the prespecified headline metric; counting \emph{both}, which the recorded criteria did not anticipate, as not correct is a deviation that changes no gate decision. Post hoc, we split \emph{both} by the name given first: \emph{new-first} is correct plus both with the new name first, and \emph{old-only} is outdated. The benchmark-track scorer also parses MCQ letters, labels unparsable answers and abstentions \emph{invalid}, and counts \emph{Thừa Thiên Huế} (renamed Huế on January 1, 2025) as outdated, which the probing-track scorer does not (kept units only). For short answers, it scores the whole first line, else an address segment by segment (a three-level one is outdated) and any other answer by the clause after a relation verb (unless negated), the first segment, or any name outside past clauses; one whose scored part names both the gold and the old unit is wrong, not \emph{both}. Post hoc rules read a lone MCQ letter, name, or address on the last line, changing 15 labels of Haiku~5.5, 610 of Sonnet~5.5, and 69 of two small models.

\subsection{Uncertainty}
\label{sec:stats}

Absorbed wards of one old province share their old and new province, so unit-level intervals are too narrow~\cite{miller2024adding}. Every interval is a paired cluster bootstrap~\cite{field2007bootstrapping} over old provinces ($B=10{,}000$; 24 or 29 provinces for absorbed units, 22 to 34 for kept units, and 7 to 11 for those in the 11 unchanged provinces), and an effect counts as established only when this 95\% interval excludes zero and a Rademacher wild cluster bootstrap test gives $p<0.05$~\cite{cameron2015practitioners}. Both postdate the fixed criteria: the prespecified criteria were evaluated with a unit-level paired bootstrap ($B=2{,}000$), and clustering changes no prespecified decision of the \nKnowers{} knowers. Over their reported runs, it removes the support of 10 cue effects established at the unit level (9 on kept units), and 18 more fail the wild test. At 0\% or 100\%, resampling intervals collapse~\cite{bowyer2025position}; 0/80 is then compatible with true rates up to 14.2\% (one-sided 97.5\% Clopper--Pearson bound, treating the 24 provinces as the effective sample size). Intervals are not corrected for multiple comparisons, and by our power calculation (after \citealp{card2020little}, with a design effect of about 2), the 80-unit sets detect only differences of roughly 20 points or more.

\section{Results}
\label{sec:results}

Bracketed intervals are province-clustered 95\% intervals in percentage points (added post hoc; Section~\ref{sec:stats}), and models run in the reasoning setting of Section~\ref{sec:inference} unless stated otherwise. Per-model tables with unit-level and clustered intervals are released.

\subsection{Few models know the reform}
\label{sec:knows}

The prespecified knowledge gate (Section~\ref{sec:prereg}) leaves \nKnowers{} knowers among the 67 distinct models (Table~\ref{tab:models}); Fig.~\ref{fig:cue} names them. Only one of the \nLocal{} local open-weight models, Muse-Glimmer-30B, passes. The others, up to 35B parameters, rarely give the new province under any cue (strict accuracy on absorbed \qt{T1} units at most 5\%). Three OpenRouter models below the gate answered only its 80 items (\mbox{Mistral-Large-4}, Gemini-4-Argon, and LongCat-2.0), so their no-cue rates and premise leak are unknown. Three runs are invalid (Section~\ref{sec:prereg}; $^*$ in Table~\ref{tab:models}) and count neither as knowers nor as non-knowers, although two of them pass the gate despite their empty outputs (Muse-Spark-1.2, 97\%; Qwen3.8-Flash, 56\%), so the knower count is a lower bound. Closed-book on all 7,995 benchmark items (Sections~\ref{sec:tracks}--\ref{sec:inference}; single runs, cues pooled; Claude with larger answer caps), the seven small models answer \mbox{32--35\%} correctly (49--53\% outdated) and the five benchmark-track knowers 37--65\% (8--47\% outdated).

\textbf{Depth.}\label{sec:depth} Models name the new province far more often than the new ward (descriptive). Of the knowers with \qt{T5} runs, GPT-6, Claude Opus~5.5, and Fable~5.1 name the successor of all 29 absorbed provinces (\qt{T6}) with no cue but the new ward (\qt{T5}) far less often (reform cue: Opus 65\%, Fable 46\%, GPT-6 40\%). DeepSeek, GLM-5.3, Claude Haiku~5.5, and Muse-Glimmer know the new province of 68--84\% of absorbed wards but the new ward of at most 4\% of renamed wards. Empty answers (up to 45\%, Claude Fable~5.1) make the \qt{T5} accuracy of reasoning models a lower bound.

\subsection{Knowers default to the pre-reform map}
\label{sec:cue}

\begin{figure}[t]
\centering
\lrefig{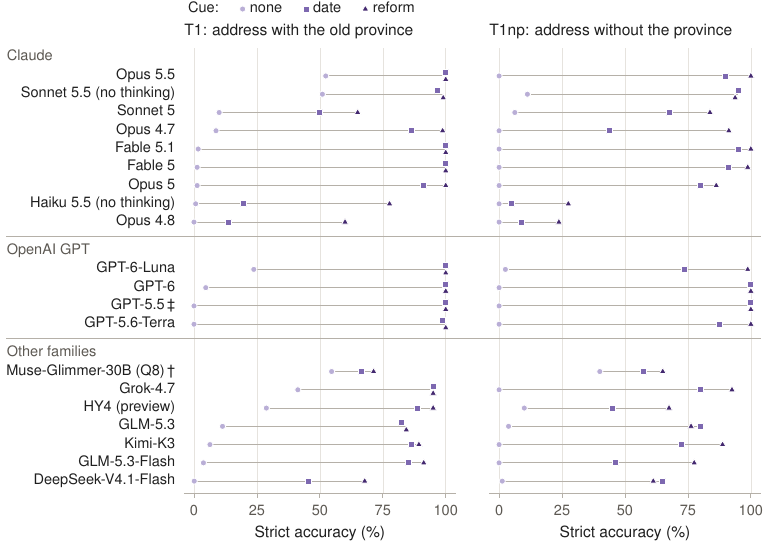}{\textwidth}
\caption{Knowers rarely give the new province unprompted; a date alone recovers it for most. Strict accuracy (\%) on absorbed units, \nKnowers{} knowers by family, sorted by \qt{T1} accuracy with no cue; left, \qt{T1}; right, \qt{T1np}. $n=80$ to 1,000 units (\qt{T1}), 80 (\qt{T1np}); no intervals shown; coinciding marks are offset. $^\dagger$The chat template supplies the date. $^\ddagger$On this route the model knew today's date (Section~\ref{sec:inference}).}
\label{fig:cue}
\end{figure}

\begin{figure}[t]
\centering
\lrefig{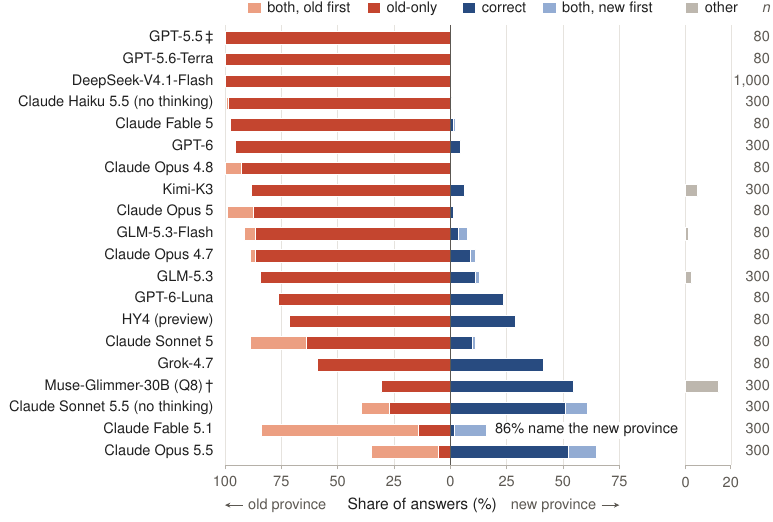}{\textwidth}
\caption{With no cue, most errors are silent old-only answers. Share of answers (\%) on absorbed \qt{T1} units with no cue, sorted by the old-only rate. Light tints: both names, on the side of the name given first (post hoc). $n$: units; no intervals shown. $^\dagger$The chat template supplies the date. $^\ddagger$On this route the model knew today's date (Section~\ref{sec:inference}).}
\label{fig:labels}
\end{figure}

With no cue (Fig.~\ref{fig:cue}), 14 of the \nKnowers{} knowers give only the new province for at most 11\% of absorbed \qt{T1} wards. At most 16\% of their answers name it first (post hoc split), and 14--100\% are old-only (Fig.~\ref{fig:labels}). GPT-5.5, whose no-cue condition is not date-free (Section~\ref{sec:inference}), still answers old-only throughout. The other six sometimes answer the new province unprompted: Muse-Glimmer (55\%; its template supplies the date), Claude Opus~5.5 (52\%), Sonnet~5.5 (51\%), Grok-4.7 (41\%), HY4 (29\%), and GPT-6-Luna (24\%). On \qt{T1np}, 19 of the 20 knowers stay at or below 11\%. Either cue recovers the update: eight knowers reach 100\% under the reform cue and six under the date cue. The two cues are within 13 points of each other except for four weaker knowers (date vs. reform cue: DeepSeek, 45 vs. 68\%; Claude Haiku~5.5, 20 vs. 78\%; Opus~4.8, 14 vs. 60\%; Sonnet~5, 50 vs. 65\%). Under the date cue, new-first rates lie within 9 points of strict accuracy for all 20 knowers, and old-only answers fall to at most 19\%, except for DeepSeek-V4.1-Flash (54\%) and Claude Haiku~5.5 (62\%).

The prespecified cue-dependence criterion (unit-level paired bootstrap) holds for all 19 API knowers: the gain of the reform cue over no cue, $\Delta_{\text{reform}}$, is $+48$ to $+100$ points, and their cluster intervals all start at $+38$ or higher. With $\Delta_{\text{date}}$, the gain of the date cue (which names no change), substituted after the criteria were fixed, every API knower reaches the 30-point threshold except Claude Haiku~5.5 ($+19$ [11,~28]) and Opus~4.8 ($+14$ points), which gain $+29$ and $+39$ on the 80 base units with the plain system prompt (Section~\ref{sec:anchor}; $+15$ and $+14$ without it). The old-only rate of Opus~4.8 still falls from 92\% to 10\%, because 72\% of its date-cue answers name both provinces, the old one first, in hedged explanations that end on the new one. Muse-Glimmer with its default template fails the criterion ($\Delta_{\text{reform}}=+17$ [8, 26], $\Delta_{\text{date}}=+12$ [3, 21]) because the template supplies the date; without that date it is cue-dependent (post hoc; $+61$ [46, 76]; Section~\ref{sec:template-res}).

\textbf{Premise-leak controls.} On absorbed units, most non-knowers do not move under any cue. Three do move under the reform cue (MiniMax-M3 $0\rightarrow43\%$, MiMo-v2.6-Flash $0\rightarrow21\%$, MiMo-v2.6-Pro $0\rightarrow19\%$) and fail the prespecified premise-leak criterion ($<0.15$), consistent with guessing a plausible merger partner once a merger is named. Under the date cue, which names no change, they stay at 1--8\%. We therefore treat $\Delta_{\text{date}}$ as the main estimate of the default and $\Delta_{\text{reform}}$ as an upper bound.

\textbf{Kept units.} Under the prespecified kept-unit rule, the cue effects of 17 of the 41 valid non-knowers with kept data are partly premise-induced: the reform cue changes their kept units by 10 points or more, mostly on \qt{T6}. Neither cue moves kept \qt{T1} and \qt{T6} units by over 3 points for DeepSeek, GPT-6, GPT-5.5, and the Claude Sonnet, Opus, and Fable knowers. Kimi-K3 and GLM-5.3 move kept wards to a neighboring province under both cues ($-18$ to $-21$ points on \qt{T1}, $-22$ to $-28$ on \qt{T1np}, all established). Muse-Glimmer does so under the reform cue ($-12$ [$-23$, $-2$]). On the \qt{T5} control (post hoc), the date and reform cues lower the accuracy by 20 points or more for Claude Sonnet~5.5 ($-22$ and $-33$), Haiku~5.5 with thinking ($-21$, $-61$; without it $-2$, $-6$), and Muse-Glimmer ($-20$, $-25$). Claude Haiku~5.5 lies near the prespecified threshold without thinking ($-9$ [$-17$, $-3$] on \qt{T1}) and fails it with thinking on \qt{T1np} ($-10$ [$-24$, $0$] under the reform cue; $-18$ [$-33$, $-6$] under the date cue). The kept-unit criterion uses point estimates on 34 to 337 units from 22 to 34 provinces, so decisions near $\pm10$ points are uncertain. Six families have a model that passes every control (DeepSeek, Zhipu, Anthropic, OpenAI, Tencent, and xAI), which meets the generality criterion; the Moonshot model (Kimi-K3) and the Meta model (Muse-Glimmer without its template date) are cue-dependent but fail the kept-unit control.

\textbf{Dual answers (post hoc).} Outside the Claude family, almost no answer names both provinces (Fig.~\ref{fig:labels}; GLM-5.3 2\%, GLM-5.3-Flash 9\%). Claude models often do: Fable~5.1 answers the old province and appends the change in 70\% of cases (2\% strict accuracy, but 86\% of its answers mention the new province), whereas Opus~5.5 gives the new province alone or first in 65\% (52\% strict) and names both, old first, in 30\%.

\textbf{Generations.}\label{sec:lineage} Of 13 Claude releases (descriptive), the four oldest do not know the reform (at most 5\% of absorbed \qt{T1} units under any cue); the nine later ones do but give at most 10\% on \qt{T1} with no cue, except Sonnet~5.5 and Opus~5.5 (51--52\%).

\textbf{Wording (post hoc).}\label{sec:np} Without the old province in the question (\qt{T1np}), old-only answers remain the rule with no cue (DeepSeek 98\%), so the old province is recalled rather than copied. The reform cue still helps every knower, by $+24$ to $+100$ points, and every cluster interval excludes zero (lowest interval: Claude Opus~4.8, $+24$~[12,~35]). Without the brevity suffix (three models, plain system prompt), answers grow from a median of 8--9 to 50--61 characters, and old-only answers stay at 95--100\%.

\textbf{Frame (post hoc).}\label{sec:frame} Asked for the province of a ward by its post-reform name, with no abolished district in the question, DeepSeek, Claude Haiku~5.5, and \mbox{Muse-Glimmer} without its template date still give the old province for 82\%, 49\%, and 57\% of the 120 names that also existed before the reform (Muse-Glimmer with its template: 23\%). Claude Opus~5.5 and Fable~5.1 instead answer the new province first (85--93\%, the old one alone for at most 1\%) but follow the old map for an old address (85--96\% old-only on \qt{T1np}): for these models, the default follows the question's frame. On \qt{T1np}, a content-free prefix adds at most 5 points to strict accuracy and 8 to the new-first rate, whereas ``hiện nay'' (\emph{currently}) matches the date for Claude Sonnet~5.5, Opus~5.5, and Fable~5.1 (92--100\% new-first), so there the date cue works as a temporal anchor. On \qt{T1}, the same prefix raises Sonnet~5.5 from 58\% to 78\% new-first and Opus~5.5 from 65\% to 92\%, so part of that cue effect is not temporal.

\textbf{Reasoning (post hoc).}\label{sec:reasoning} The ladder holds without reasoning (DeepSeek-V4.1-Flash, thinking off, gains $+68$ [52, 83] under the reform cue). Turning on thinking does not remove the default, and for Claude Sonnet~5.5 it lowers no-cue accuracy (51\% without thinking vs.\ 9\% with it, when 42\% name both, old first), although it helps a weak knower apply a cue (Claude Haiku~5.5 under the date cue: 20 $\rightarrow$ 94\%). Claude Fable~5.1 gives the same picture at its default setting as at the lowest (\qt{T1}, no cue: 1 vs.\ 2\% strict, 24 vs.\ 14\% old-only; reasoning on 79 vs.\ 82\% of cue-ladder items).

\subsection{Date placement changes which map is used (post hoc)}
\label{sec:anchor-res}

\begin{figure}[t]
\centering
\lrefig{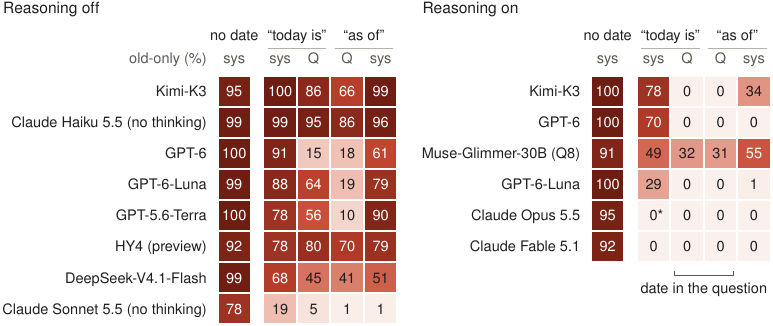}{\textwidth}
\caption{No placement leaves fewer old-only answers than the better of the two in-question dates; with reasoning, a date in the question leaves none except for Muse-Glimmer. Old-only answers (\%; darker vermilion, more) on the 80 absorbed \qt{T1np} units with no cue. Columns: plain system prompt; ``today is'' in the system prompt (sys) or the question (Q); ``as of'' as the prefix of Q or an instruction in sys (Section~\ref{sec:anchor}). Rows: by reasoning and old-only rate, both in the ``today is'' sys column. No intervals shown. $0^\ast$: Claude Opus~5.5 names both provinces, old first, in 96\% of its answers.}
\label{fig:placement}
\end{figure}

\textbf{System prompt or question.}\label{sec:placement-res} With a plain system prompt, every model in Fig.~\ref{fig:placement} gives 78--100\% old-only answers. Today's date in the system prompt (the ``today is'' sentence) leaves most answers old-only with reasoning off: of the 12 models (the figure's eight and Claude Opus~4.7, Opus~4.8, Opus~5, and Sonnet~5, not shown because they lack one cell; reasoned on at most 18\% of these items), 10 give 68--100\% old-only and 0--30\% new-first answers; only Claude Sonnet~5.5 (19\%) and Opus~4.7 (31\%) give fewer. With reasoning, the same date leaves 70--78\% old-only answers for GPT-6 and Kimi-K3 and 29--49\% for GPT-6-Luna and Muse-Glimmer. Fable~5.1 answers the new province first (90\%), and Opus~5.5 names both, almost always the old province first (96\%; as in Fig.~\ref{fig:protocol}, right), which a parser that takes the first name reads as outdated.

The date works better in the question. Without reasoning, the ``as of'' prefix in the question lowers the old-only rate relative to today's date in the system prompt for all eight models in the figure (GPT-6 $-74$ [$-83$, $-66$]; established for all but HY4, $-8$ [$-22$, $+8$]), and raises the new-first rate to 14--96\%. Still, 10--86\% of the answers stay old-only for every model except Sonnet (1\%). The same ``today is'' sentence moved into the question helps six of the eight ($-14$ to $-76$, all established), whereas the ``as of'' instruction in the system prompt helps only three and makes GPT-5.6-Terra worse ($+12$ [4, 22]). With reasoning, ``as of'' in the question leaves no old-only answer for seven of eight models (80--100\% new-first; Kimi-K3 $-66$ [$-80$, $-50$] relative to reasoning off); the exception is Muse-Glimmer (31\%). The seven include GLM-5.3 (80\% new-first) and Claude Fable~5 (100\%), which ran only some placements and are not in Fig.~\ref{fig:placement}; GPT-5.6-Terra, asked to reason, did so on 1--36\% of items in three of five conditions, and HY4's reasoning runs are invalid in two. On 300 units from all 29 absorbed provinces (six models, reasoning off), ``as of'' in the question again lowers the old-only rate for every model ($-12$ [$-21$, $-3$] for HY4 to $-76$ [$-82$, $-70$] for GPT-6).

Told in the system prompt that today is January 1, 2025, every model gives 80--100\% old-only answers on absorbed \qt{T1} wards, including Claude Opus~5.5, which gives 6\% old-only answers on the same 80 \qt{T1} units under the plain system prompt (Fig.~\ref{fig:placement} shows \qt{T1np}). The post-reform date in the same position leaves 41--99\% old-only answers for nine of the 12 models with reasoning off (Claude Opus~4.7, 14\%; Opus~5 and Sonnet~5.5, 1\%). In the question, ``as of January 2025'' gives 88--100\% old-only answers for nine of 11 models tested but 45--46\% for GPT-6-Luna and HY4.

\textbf{Agent settings and chat template.}\label{sec:agent-res}\label{sec:template-res} Offered a \texttt{get\_current\_date} tool without a call instruction, five of six models without reasoning never called it, and \mbox{GPT-6-Luna} called it for 29\% of the questions, each time answering with the old province. With the tool result or an earlier turn (reasoning off), 69--99\% of answers stay old-only (\mbox{1--28\%} new-first), and none of these conditions moves a kept ward. Only with reasoning on and an instruction to call the tool does GPT-6-Luna mostly apply the fetched date (29\% old-only). Rewriting the question with DeepSeek lowers the old-only rate for all six answering models but inserts the old province as a premise into 7 of the 80 absorbed questions (9\%) and costs 8--12 points on kept units. In Muse-Glimmer's chat template, the system block stating a knowledge cutoff and today's date switches the map: without it the model shows the API models' default (on \qt{T1}, 9\% correct and 89\% old-only), and with a pre-reform date it gives 99\% old-only answers.

\subsection{Retrieval and the \texorpdfstring{$2{\times}2$}{2\texttimes 2} control}
\label{sec:rag}

\begin{figure}[t]
\centering
\lrefig{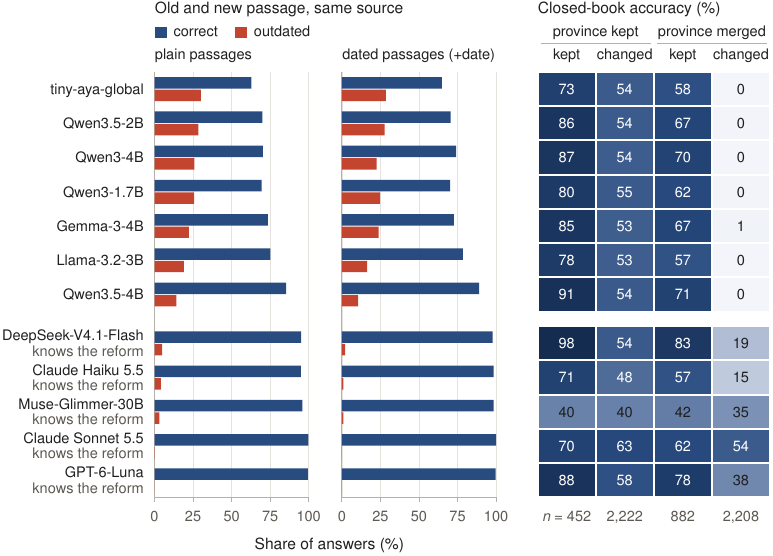}{\textwidth}
\caption{Seven small models unaware of the reform give outdated answers in 14--30\% of cases with plain passages, the five that know it in 0--5\%; closed-book, the seven answer 0--1\% of changed wards in merged provinces correctly. Left: same-source passages, plain or dated (+date), 1,262-item subset, $n=390$ per bar. Right: closed-book accuracy (\%; darker navy, higher) per $2{\times}2$ cell (header rows: province, then ward) on the \qt{T1} membership, \qt{T4}, and \qt{T5} items of the full set, without the 779 copyable items; $n$ below (Fig.~\ref{fig:reform} also counts \qt{T1} new-name, \qt{T3}, and copyable items). Claude rows: no thinking; larger answer caps (Section~\ref{sec:inference}). No intervals shown.}
\label{fig:rag-cells}
\end{figure}

\textbf{Retrieval.} Figure~\ref{fig:rag-cells} summarizes the benchmark track (descriptive; single runs of 12 models; cues pooled). On the stratified 1,262-item subset, a post-reform passage lets the small models answer 71--98\% of the items correctly, but with old and new passages both from Wikipedia or both templates, 14--30\% of their answers are outdated (Fig.~\ref{fig:rag-cells}, left), against 2--19\% with the new passage alone, and stating each passage's date recovers at most 4 points. The five knowers, which also differ from the small models in size or provider, give outdated answers in only 0--5\% (0--0.8\% with the new passage alone). With a new Wikipedia excerpt and an old template passage, the small models give outdated answers to 32--62\% of the 220 such items of the full set (1--34\% with the new passage alone), against 1--16\% for the knowers (Haiku 16\%, Muse-Glimmer 10\%, the others 1--6\%). Without diacritics, \qt{T4} accuracy with the post-reform passage (400 twin pairs) changes by at most 6 points for six of the seven small models; Qwen3.5-2B rises from 5\% to 49\%, because with diacritics it mostly gives the old address (74\%).

\textbf{The $2{\times}2$ control.} The small models answer 73--91\% of the pure control ($\pc{}=0,\wc{}=0$) correctly but 0--1\% of the cell $\pc{}=1,\wc{}=1$ (province merged, ward changed), where 88--95\% of their answers are pre-reform (Fig.~\ref{fig:rag-cells}, right). Because each pure-control item names the unit it asks about, this shows that they keep a named unit that did not change, not that they know its geography. In the two middle cells, 54\% and 76\% of the items ask for an unchanged name, so these cells isolate no province or ward effect. Muse-Glimmer answers only 40\% of the pure control correctly (other cells 35--42\%); it names another successor for 80\% of its pure-control \qt{T5} items (the other knowers for 4--44\%). DeepSeek, \mbox{GPT-6-Luna}, Haiku~5.5, and Sonnet~5.5 answer 19\%, 38\%, 15\%, and 54\% of the cell $\pc{}=1,\wc{}=1$ correctly (70\%, 31\%, 51\%, and 13\% outdated), but only 0\%, 4\%, 1\%, and 37\% of its cue-free \qt{T1} and \qt{T5} items (DeepSeek and GPT-6-Luna: 621 and 566 of 624 outdated); on these types, the reform cue lowers Haiku's pure control from 99\% to 33\%, and either cue lowers \mbox{GPT-6}-Luna's from 97\% to 81--82\%. Sonnet's 70\% (pure control) and 54\% would be about 54\% and 47\% if counted as in Section~\ref{sec:inference}, and 63\% and 51\% without the post hoc last-line rules.

\section{Discussion}
\label{sec:discussion}

\subsection{What the default means}
\label{sec:disc-default}

A closed-book probe that asks plainly would report most knowers as outdated, mistaking a default for missing knowledge, and one that names the event would report them as updated, overstating what a user typing an old address receives. We do not measure training data. The old map may dominate because effective cutoffs often lag reported ones~\cite{cheng2024dated} and older text about it abounds.

\subsection{How a date takes effect}
\label{sec:disc-date}

A date in the system prompt may set the present, and one in the question may scope the query. Reasoning changes how a date is used, but the default persists on items without measured reasoning, though short internal reasoning may go unmeasured.

Prompted earlier cutoffs suppress knowledge causally related to a later event in only about 19\% of cases for models without reasoning~\cite{gao2025rewind}. In our runs, a pre-reform date in the system prompt gives 80--100\% old-only answers on absorbed \qt{T1} units for every model tested, but 12 of these 16 models give 89--100\% already without a date. The other four (Claude Opus~5.5, Fable~5.1, Sonnet~5.5, and Muse-Glimmer) also respond to a post-reform date there (6--70\% to 0--38\% old-only), so these runs do not show that the two directions of a date differ (Section~\ref{sec:placement-res}).

\subsection{Using the resource}
\label{sec:disc-use}
\label{sec:disc-implications}

Evaluations on \textsc{ViAdminShift} should report new-first and old-only rates alongside strict accuracy, report whether the model actually reasoned (as measured, not as requested), include the kept units, and cluster intervals by old province. They should exclude the 779 copyable items (listed in the release) from per-type and $2{\times}2$ analyses (middle cells: Section~\ref{sec:rag}), and the 19 \qt{T3} items on new wards from outdated rates.

In our runs, scoping the question in time (``as of $\langle$date$\rangle$'') worked better than today's date in the system prompt or a date tool, so address-handling services may prefer it; they should also check unchanged units, because the date cue makes Kimi-K3 and GLM-5.3 move kept wards to a neighboring province (Section~\ref{sec:cue}). Operators should check the deployed chat template, which may insert a date they never see, and probe each endpoint for a date added by the provider~\cite{sanzguerrero2026dating}. Dual answers are safer than silent old ones only if parsers do not take the first name.

\subsection{Limitations}
\label{sec:limitations}

\begin{itemize}
\item \textbf{Scope.} One reform, one language, and our templates, not user queries: the design transfers to other reorganizations; our numbers may not. A question naming an abolished district also admits the old answer as a literal reading; the post-reform names of Section~\ref{sec:frame} suggest this reading for Claude Opus~5.5 and Fable~5.1, not for DeepSeek and Haiku~5.5. In the cell $\pc{}=1,\wc{}=1$, \qt{T1} membership, \qt{T4}, and \qt{T5} items cover only wards whose province name changed, the control cells are small (43 old wards in the pure control, 84 in the other), and the \cue{date} cue uses one date.
\item \textbf{Data.} Gold answers follow the structure of July 1, 2025, not the 2026 unit-type changes, which precede our prompt date (Section~\ref{sec:release}). Divided wards appear in no item whose answer is a new ward, and the five island special zones in no ward-level item. The NSO table determines every pair but one (only two independent lineages exist), manual checks are small samples, and the fine structure of divided wards was hand-checked for four of the 441. Public items may enter later training data.
\item \textbf{Models and the gate.} The gate is behavioral: stated cutoffs are rarely available, so whether they predict it could not be checked. Below it, the reform-cue gains of MiniMax-M3 and the two MiMo models (19 to 43 points) may be premise-induced guessing; above it, a model may have seen only coverage of the reform, and accuracies near 0.50 may be partly premise-induced (non-knowers reach 43\%). The local-model evidence, including the chat-template experiment, rests on \mbox{Muse-Glimmer}-30B, run at 8 bits with no 16-bit comparison.
\item \textbf{Runs and endpoints.} In three repeated runs of the 80 absorbed \qt{T1} units for four models, an item gets the same label in all three runs for 80--100\% of the items of DeepSeek-V4.1-Flash and Muse-Glimmer (no template date), but for only 56\% and 59\% in two cells with intermediate rates (Claude Sonnet~5.5 at \cue{none}, Haiku~5.5 at \cue{date}) and 74\% for Haiku~5.5 at \cue{reform}, so single-run rates are not properties of items; aggregate rates varied by up to 10 points across repeats, and adding the plain system prompt moved single cells by up to 45 (Section~\ref{sec:anchor}); the intervals include neither. Ward-level results rest on 24 to 29 province mappings (Section~\ref{sec:stats}). API results reflect the models as served, with no record of provider or snapshot, and no-date conditions off the probed reseller may contain a provider-added date (Section~\ref{sec:inference}).
\item \textbf{Date experiments.} GLM-5.3 ran only with ``today is'' in the system prompt and ``as of'' in the question, confounding placement and phrasing. The agent settings use one tool and one system instruction, with reasoning only for GPT-6-Luna.
\item \textbf{Benchmark track.} It used seven small models that do not know the reform and five that do, Muse-Glimmer-30B and four API models (Section~\ref{sec:tracks}), which also differ from them in size or provider; Claude closed-book results depend on answer caps (Sonnet's most), and most evidence on knowers comes from the probing track. For 79\% of the items (6,303 of 7,995), the new passage is a template, cleaner than real documents, so retrieval results likely overstate performance on mixed-date text.
\item \textbf{Scoring.} The probing-track scorer was validated only on 200 province answers (\qt{T1}, \qt{T1np}), by the two authors, blind but knowing the rules; it undercounts new-first answers. The benchmark-track scorer is not validated, and its last-line rules and the refinements made after all runs (datasheet) are post hoc.
\end{itemize}

\section{Conclusion}
\label{sec:conclusion}

\textsc{ViAdminShift} turns the 2025 reform into a Vietnamese test of updated knowledge. Its old-to-new tables, checked against an independent lineage, yield 7,995 items with exact post-reform answers, a $2{\times}2$ control of unchanged units, diacritic-free address twins, and old and new retrieval passages. Most knowers default to the pre-reform map, and placing the question in time usually switches them to the new one. Post hoc, a date in the question works best, whereas with reasoning off, today's date in the system prompt, a tool result, or an earlier turn leaves most answers silently outdated. Small models without the update often give outdated answers with mixed passages; the five benchmark-track knowers, which also differ in size or provider, rarely do.

Future work can test other event-level updates and languages, and whether fine-tuning or knowledge editing can move a model's default, not only its knowledge.

\section*{Ethics and Data Statement}

We use the data statement schema~\cite{bender2018data} for a written, template-generated resource; the released datasheet answers the other questions.

\textbf{Curation rationale.} Official tables fix thousands of facts changed on one date. Items cover all 63 old and 34 new provinces and 1,708 old wards in all four $2{\times}2$ cells.

\textbf{Language variety.} All text is standard written Vietnamese (vi-VN, Unicode NFC) with official names; only the 400 \qt{T4} twins lack diacritics (Section~\ref{sec:vietnamese}).

\textbf{Speakers, annotators, and text.} Except for the Wikipedia editors (demographics unknown), no speakers or users are represented. No crowd workers were employed; the manual checks and the scorer-validation labels were done by the two authors, both native speakers (Section~\ref{sec:qc}). Items are short, well-formed questions about one unit, so results may not transfer to real queries that drop diacritics, abbreviate names, or mix old and new names. The data contain no personal information; \qt{T4} addresses use synthetic house numbers and hamlets.

\textbf{Speech situation.} All text is written and edited: template questions and passages over official unit names, meant for evaluating models, and Wikipedia revisions from before and after the reform (Section~\ref{sec:contexts}), written for general readers.

\textbf{Sources and licenses.} Wikipedia text is under CC~BY-SA (4.0 since June 2023; earlier 3.0 text may be adapted under 4.0), and each passage carries its page title and revision ID. The NSO file, its verbatim copy in VietnamProvinces~\cite[][GPL-3.0-or-later]{vietnamprovinces}, the data files of vietnamadminunits (MIT), undertheseanlp/\allowbreak{}address (Apache-2.0), and vietaddr (GPL-3.0-or-later), and the HTML reproductions of the resolutions are downloaded by script, not redistributed (Section~\ref{sec:release}). Legal texts are not protected by copyright (Law on Intellectual Property, Art.~15). Models were used for noncommercial research under their licenses or terms of use.

\textbf{Collection, compute, and intended use.} All sources were retrieved from their public sites (Wikipedia via its API, about two requests per second), and local models ran on a rented four-GPU RTX~4090 server and on CPU servers. \textsc{ViAdminShift} measures model knowledge; it is not an address converter.

\backmatter

\section*{Statements and Declarations}

\textbf{Competing interests.} The authors have no competing interests to declare.

\textbf{Funding.} No funding was received for this work.

\textbf{Ethics approval, consent to participate, and consent for publication.} Not applicable: no human participants or personal data are involved (see the Ethics and Data Statement), and the scorer-validation labels are the authors' own.

\textbf{Data availability.} The \textsc{ViAdminShift} data (JSON Lines), version 2025-07-01, with the exact prompts, annotation sheets, and model outputs behind every result, are available at \url{https://github.com/LLaammTTeerr/ViAdminShift} and archived at \url{https://doi.org/10.5281/zenodo.23473403}; Section~\ref{sec:release} and the datasheet describe them and their licenses. Source files with NSO or government-portal data are not redistributed, nor are the canonical tables.

\textbf{Code availability.} All code is in the same release (MIT license).

\textbf{Author contributions.} Phan Binh Nguyen Lam: conceptualization, methodology, software, data curation, formal analysis, investigation, validation, visualization, and writing (original draft). Nguyen Thai Anh: conceptualization, methodology, validation, supervision, and writing (review and editing). Both authors read and approved the final manuscript.

\textbf{Use of large language models.} The authors designed the study, made the manual checks and the scorer-validation labels themselves, checked the data and the results, and take full responsibility for the content of this paper. Claude Opus~5.5 (Anthropic) was used as a coding and writing assistant: it helped write the code and helped draft and edit the text. Claude models are also among the evaluated systems.

\bibliography{references}

\end{document}